\documentclass{article}

\usepackage{microtype}
\usepackage{graphicx}
\usepackage{subcaption}
\usepackage{booktabs} 

\usepackage{hyperref}

\usepackage[preprint]{icml2026}

\usepackage{amsmath}
\usepackage{amssymb}
\usepackage{mathtools}
\usepackage{amsthm}

\usepackage{colortbl}
\usepackage{multirow}
\usepackage{xcolor}
\usepackage{soul}
\usepackage{array}
\usepackage{seqsplit}

\usepackage[capitalize,noabbrev]{cleveref}

\theoremstyle{plain}

\theoremstyle{definition}

\theoremstyle{remark}

\usepackage[textsize=tiny]{todonotes}

\icmltitlerunning{Data Distribution Matters: A Data-Centric Perspective on Context Compression for Large Language Model}

\begin{document}

\twocolumn[
  \icmltitle{Data Distribution Matters: A Data-Centric Perspective on \\ Context Compression for Large Language Model}


  \icmlsetsymbol{equal}{*}

  \begin{icmlauthorlist}
    \icmlauthor{Kangtao Lv}{equal,comp,xxx}
    \icmlauthor{Jiwei Tang}{equal,yyy}
    \icmlauthor{Langming Liu}{equal,comp}
    \icmlauthor{Haibin Chen}{comp}
    \icmlauthor{Weidong Zhang}{comp}
    \icmlauthor{Shilei Liu}{comp}
    \icmlauthor{Yongwei Wang}{xxx}
    \icmlauthor{Yujin Yuan}{comp}
    \icmlauthor{Wenbo Su}{comp}
    \icmlauthor{Bo Zheng}{comp} 
  \end{icmlauthorlist}

  \icmlaffiliation{comp}{Future Living Lab of Alibaba}
  \icmlaffiliation{yyy}{Tsinghua University}
  \icmlaffiliation{xxx}{Zhejiang University}

  \icmlcorrespondingauthor{Bo Zheng}{bozheng@alibaba-inc.com}

  \vskip 0.3in
]




\printAffiliationsAndNotice{\icmlEqualContribution}
\begin{abstract}
The deployment of Large Language Models (LLMs) in long-context scenarios is hindered by computational inefficiency and significant information redundancy. Although recent advancements have widely adopted context compression to address these challenges, existing research only focus on model-side improvements, the impact of the data distribution itself on context compression remains largely unexplored. To bridge this gap, we are the first to adopt a \textbf{data-centric} perspective to systematically investigate how data distribution impacts compression quality, including two dimensions: \textbf{input data} and \textbf{intrinsic data} (\textit{i.e.}, the model's internal pretrained knowledge). We evaluate the semantic integrity of compressed representations using an autoencoder-based framework to systematically investigate it. Our experimental results reveal that: (1) encoder-measured input entropy negatively correlates with compression quality, while decoder-measured entropy shows no significant relationship under a frozen-decoder setting; and (2) the gap between intrinsic data of the encoder and decoder significantly diminishes compression gains, which is hard to mitigate. Based on these findings, we further present practical guidelines to optimize compression gains.
\end{abstract}

\section{Introduction}
Large Language Models (LLMs) have become foundational infrastructure in Natural Language Processing (NLP) due to their exceptional language modeling and generalization capabilities~\cite{qwen2025qwen25technicalreport,team2025kimi,liu2025deepseek,zeng2025glm,DBLP:journals/corr/abs-2504-07282,DBLP:journals/corr/abs-2511-12913,DBLP:conf/kdd/LiuLY0YZWLWS00025}. However, in practical applications such as Retrieval-Augmented Generation (RAG)~\cite{lewis2020retrieval}, In-Context Learning (ICL)~\cite{dong-etal-2024-survey}, or large-scale code repository analysis, LLMs often need to process input sequences with even tens of thousands of tokens. This requirement exposes two bottlenecks: (1) The self-attention mechanism in Transformer architectures incurs quadratic time complexity with respect to sequence length, leading to sharply increased inference latency and computational cost~\cite{vaswani2017attention,ge2024incontext,tang-etal-2025-perception}. (2) Semantic redundancy commonly present in long texts not only dilutes the density of key information but also introduces noise, thereby degrading performance on downstream tasks~\cite{jiang-etal-2024-longllmlingua,liu2024forgettingcurvereliablemethod,liu-etal-2024-lost,tang2025gmsa}.

\begin{figure}
    \centering
    \includegraphics[width=1\linewidth]{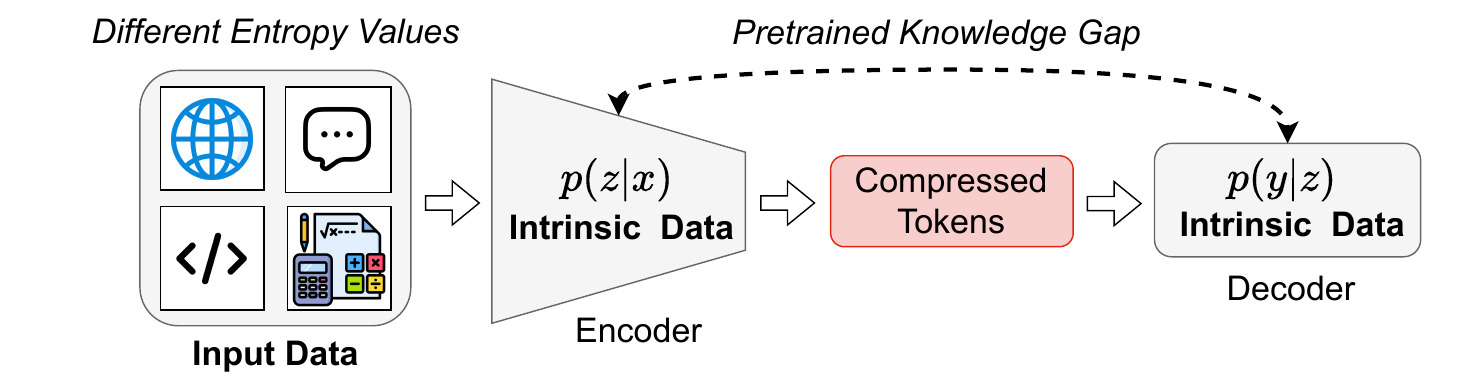}
    \caption{A illustration of an encoder-decoder architecture for context compression. Input data such as web pages, text, or code is mapped by the encoder into a latent representation $p(z|x)$, where different types of input data exhibit varying entropy values. The compressed tokens are then fed through the decoder to produce an output distribution $p(y|z)$. The dashed arrow denotes the intrinsic data gap between the encoder and the decoder. Our goal is to investigate, from a data-centric perspective, how input data with \emph{various entropy values} and the intrinsic data (\textit{i.e.}, \emph{the model’s internal pretrained knowledge}) gap impact the compression quality.}
    \label{fig:intro}
\end{figure}

To address these challenges, context compression methods~\cite{jiang-etal-2024-longllmlingua,pan-etal-2024-llmlingua,tang-etal-2025-perception,zhou2025mooscomp,cao2025efpcefficientflexibleprompt,Zhao_Wu_Xu_2025,chirkova2025provence,hwang-etal-2025-exit,cui2025core,DBLP:journals/corr/abs-2509-15763,DBLP:conf/iclr/Zhang0XSYD25} compress the original long context into a compact set of tokens, aiming to drastically reduce input sequence length and semantic redundancy. These methods have achieved substantial progress primarily through \textbf{model-side} improvements (e.g., diverse compression strategies and architectural modifications). However, evidence across multiple areas suggests that data distribution can have a larger effect on performance than model architecture or training strategy \cite{sorscher2022beyond,hoffmann2022training,zhou2023lima,zha2023data,zha2025data}.
In context compression, this \textbf{data-side} factor remains underexplored (Figure~\ref{fig:intro}). Neglecting it not only obscures the sources of variability in compression performance but also hinders reliable deployment across domains. In particular, two questions are still unclear: (1) \textit{how the distribution of input data affects compression}, and (2) \textit{how the intrinsic data (\textit{i.e.}, model's pretrained knowledge) affects compression.}

To bridge this gap, we are the first to adopt a \textbf{data-centric} perspective and systematically investigate how \textbf{input data} and \textbf{intrinsic data} (\textit{i.e.}, the model's pretrained knowledge) distributions influence the context compression quality. We use an autoencoder-based framework to evaluate semantic preservation completeness (\textit{i.e.}, compression quality), as reconstruction directly reflects information integrity, offering interpretability and enabling quantitative measurement of semantic loss. In this framework, an encoder compresses input context into learnable latent vectors, and a decoder attempts to reconstruct the original text from these vectors. Both encoder and decoder are pretrained \emph{from scratch} on data drawn from diverse distributions, ranging from web-crawled general text to logic-intensive domains such as mathematics and code. \emph{Under this paradigm, we can control the intrinsic data gap between encoder and decoder and analyze how different pretrained knowledge impact compression quality. To enable a unified measure across different types of input data, we employ information entropy~\cite{DBLP:journals/bstj/Shannon48} to characterize input data complexity and quantitatively assess its impact on compression quality.} Our experimental results demonstrate two key findings:
(1) The input data entropy quantified by the encoder exhibits a pronounced negative correlation with compression quality, whereas the entropy measured by the decoder shows no significant correlation. This indicates that the decoder's perception of complexity is heavily biased by its internal prior when the input deviates from its intrinsic distribution. (2) As the gap in intrinsic data widens, compression quality degrades significantly. Crucially, we reveal a fundamental asymmetry: the compression quality is primarily governed by the decoder's intrinsic distribution rather than the encoder's, suggesting that decoder alignment is the primary bottleneck for context compression. Both findings demonstrate that \textbf{data distribution matters}.

Our main contributions are threefold: 

(1) We identify the significant impact of data distribution on compression quality both between input data and intrinsic data. 

(2) We conduct a systematic analysis from a data-centric perspective on how data distributions influence context compression, examining the roles of input data and intrinsic data, and provide a theoretical analysis in Appendix~\ref{apx:theoretical_analysis}. 

(3) We present a set of compression guidelines, offering principled strategies to optimize compression gains.

\section{Related Work}
\textbf{Context Compression Methods.} Existing methods to context compression can be broadly categorized into two types: hard prompt compression and soft prompt compression. Hard prompt compression involves selecting a subset of important tokens from the original context~\cite{li-etal-2023-compressing,jiang-etal-2023-llmlingua,jiang-etal-2024-longllmlingua,pan-etal-2024-llmlingua,tang-etal-2025-perception,zhou2025mooscomp,cao2025efpcefficientflexibleprompt,Zhao_Wu_Xu_2025,chirkova2025provence,hwang-etal-2025-exit} or generating a summary~\cite{yoon-etal-2024-compact,xu2024recomp,cui2025core}, while soft prompt compression aims to compress long context into a significantly shorter set of implicit semantic vectors~\cite{mu2023learning,cheng2024xrag,li2024500xcompressorgeneralizedpromptcompression,ge2024incontext,tang2025gmsa,DBLP:journals/corr/abs-2509-15763,DBLP:conf/iclr/Zhang0XSYD25,zhao2025positionidsmatterenhanced,liu2025autoencodingfreecontextcompressionllms}. Although these methods effectively reduce input sequence length and achieve strong performance on various downstream tasks, they only focus on model-side improvements (\textit{e.g.}, designing novel compression strategies, modifying model architectures) while ignoring the inherent impact of data itself on compression quality. In many scenarios, data distributions exert a greater impact on model performance than architectural choices or training strategy adjustments. \emph{Ignoring data distributions not only makes it difficult to explain fluctuations in compression effectiveness but also hinders reliable deployment across domains. In this work, we adopt a data-centric perspective and systematically investigate how both input and intrinsic data distributions impact the quality of context compression.}

\begin{figure*}[ht]
    \centering
\includegraphics[width=0.8\textwidth]{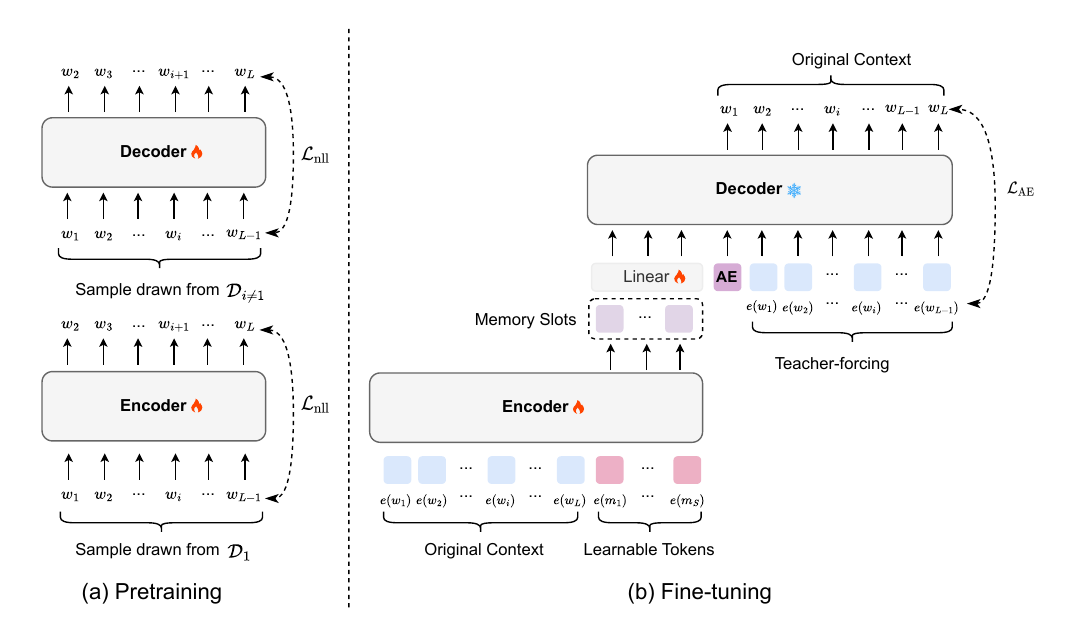}
    \caption{The overall framework. Our framework includes two phases: (a) pretraining and (b) fine-tuning. \textbf{Pretraining Phase:} We begin by independently pre-training a randomly initialized encoder and decoder. To establish an \textit{intrinsic data discrepancy} between the two modules, the encoder is pre-trained on dataset $\mathcal{D}_1$, whereas the decoder is pre-trained on a disjoint set $\mathcal{D}_{i \neq 1}$. The optimization objective for this phase is the standard negative log-likelihood loss $\mathcal{L}_{\text{nll}}$. \textbf{Fine-tuning Phase:} Following pre-training, the encoder and decoder are coupled to form a joint autoencoder architecture, where $e(\cdot)$ denote the embedding look-up function that maps tokens to their latent representations. To guide the decoder, we introduce a specialized indicator token $\texttt{[AE]}$, which prompts the decoder to perform the context reconstruction task. . During this stage, the model is trained using the autoencoder reconstruction loss $\mathcal{L}_{\text{AE}}$. Crucially, we optimize the encoder parameters while keeping the decoder \textit{frozen}.}
    \label{fig:framework}
\end{figure*}

\section{Methodology}
Prior studies on context compression \emph{only} focus on how to make compression “good,” while the impact of data remains largely unexplored. As shown in Table~\ref{tab:influence of the distribution shift}, when the endogenous data distributions of the encoder and decoder are misaligned (e.g., one is a base model\footnote{https://huggingface.co/Qwen/Qwen2.5-0.5B} and the other is a code model\footnote{https://huggingface.co/Qwen/Qwen2.5-Coder-0.5B}) performance degrades compared to the aligned setting, particularly on token-matching metrics such as BLEU. The result underscores that data materially affect compression performance. However, the proprietary opacity of pre-training corpora in models such as Qwen precludes a granular understanding of how specific data characteristics dictate performance. Motivated by this limitation, we adopt a data-centric perspective to systematically investigate how both input data and intrinsic model data influence the quality of context compression.

\begin{table}[th]
\small
\setlength{\tabcolsep}{3pt}
\centering
\caption{Autoencoding results when using Qwen2.5-Base-0.5B and Qwen2.5-Coder-0.5B as the backbone respectively, under the unified fine-tuning on dataset $\mathcal{D}_1$. All metrics are reported as percentages (\%).}
\resizebox{ \columnwidth}{!}{%
\begin{tabular}{c c | c c c}
\toprule
\textbf{Encoder} & \textbf{Decoder} & \textbf{F1}\phantom{\,{\color{red}$\downarrow$}} & \textbf{BLEU}\phantom{\,{\color{red}$\downarrow$}} & \textbf{ROUGE-L}\phantom{\,{\color{red}$\downarrow$}} \\
\midrule
Qwen2.5-Base & Qwen2.5-Base & 96.96\phantom{\,{\color{red}$\downarrow$}} & 80.02\phantom{\,{\color{red}$\downarrow$}} & 96.57\phantom{\,{\color{red}$\downarrow$}} \\
\midrule
Qwen2.5-Coder & Qwen2.5-Base & 92.15\,{\color{red}$\downarrow$} & 64.61\,{\color{red}$\downarrow$} & 90.93\,{\color{red}$\downarrow$} \\
Qwen2.5-Base & Qwen2.5-Coder & 95.38\,{\color{red}$\downarrow$} & 72.25\,{\color{red}$\downarrow$} & 94.89\,{\color{red}$\downarrow$} \\
\bottomrule
\end{tabular}
}%
\label{tab:influence of the distribution shift}
\end{table}

\subsection{Problem Formulation}
Context compression typically involves a compressor that maps a long text sequence of length $L$ into a much shorter latent representation sequence of length $S$, where $L>>S$. This can be formulated as:

\begin{equation}
\min _{\theta} \mathbb{E}_{x, y \sim \mathcal{D}} \left[ \text{KL} \left( p(y \mid x) \parallel p(y \mid E(x;\theta_E)) \right) \right] \, ,
\end{equation}
where $p(\cdot)$ is the probability distribution function of the decoder, $E(\cdot;\theta_E)$ is the encoder.

In this work, we focus on \emph{the impact of data} on context compression. To measure compression capability, we evaluate the model's ability to reconstruct the original context from the compressed representation, thereby quantitatively characterizing the integrity of the compressed information and the accuracy of the model’s memory. Specifically, we employ a typical autoencoder~\cite{kramer1991nonlinear} framework using LLMs as the backbone for both the encoder and the decoder $D(\cdot ;\theta_D)$. The process, which involves compressing and subsequently reconstructing context $x$ of length $L$, can be formulated as a transformation:
\begin{equation}
  \label{eq:compression and reconstuction}
x \xrightarrow[\text{compress}]{E(p(z|x);\theta_E)} z \xrightarrow[\text{reconstruct}]{D(p(x'|z);\theta_D)} x'.
\end{equation}
The optimization objective is to minimize the loss function $\mathcal{L}_{\text{AE}}(x,x')$. Here, the encoder serves as the compressor, producing a latent representation $z$ from the input $x$ to condense long context, while the decoder serves as the predictor that reconstructs information from $z$ to obtain $x'$.

\subsection{Data Preparation}
\label{sec:Data Preparation}
Our data is sourced from \textbf{The Pile}~\cite{gao2020pile} (more details in Appendix~\ref{apx:pile}). To systematically investigate the impact of domain-specific knowledge, we curate six distinct datasets, each containing $50$ billion tokens as processed by the Qwen3 tokenizer.
The datasets are structured as follows: \textbf{dataset $\mathcal{D}_1$} consists entirely of the Common Crawl (CC) subset, representing a baseline of general-purpose web text. \textbf{dataset $\mathcal{D}_2$ to dataset $\mathcal{D}_6$} are constructed by substituting portions of the CC data with a domain-specific mixture comprising ArXiv, GitHub, and DM Mathematics in a fixed ratio of $2:2:1$. Specifically, the proportion of this specialized mixture $\alpha \in \{1/6, 2/6, 3/6, 4/6, 5/6\}$ increases linearly across the five datasets. \textit{This experimental design is motivated by the need to investigate the divergence in compression performance between general natural language and formal logical corpora. While natural language is typically associated with information-seeking tasks, code and mathematics are emblematic of logical reasoning, representing two primary and functionally distinct categories of textual data in language modeling.}
For evaluation, we reserve 100k and 10k samples from the CC subset for fine-tuning and testing, respectively. \textbf{All data splits are strictly partitioned to ensure zero overlap.}

\subsection{Training}
To mitigate confounding effects introduced by compression algorithm itselves, we adopt the classical context compression approach ICAE~\cite{ge2024incontext} (\textbf{I}n-\textbf{C}ontext \textbf{A}uto\textbf{E}ncoder) framework. This setup is designed to rigorously evaluate the fidelity with which compressed representations retain semantic information and reconstruct them. The process involves an LLM-based encoder that encodes an input context $c=(w_1,w_2,…,w_L)$ into a small number of memory slots $(\widetilde{m}_1, \ldots, \widetilde{m}_k)$. A corresponding LLM-based decoder is then reconstructs the original context $c$ conditioned on these memory slots. The training stage can be split to pretraining and fine-tuning. 
\paragraph{Pretraining.} We independently pre-train a randomly initialized encoder and decoder on disjoint datasets $\mathcal{D}_1$ and $\mathcal{D}_{i \neq 1}$, respectively. Both modules are optimized using the standard negative log-likelihood loss $\mathcal{L}_{\text{nll}}$. Specifically, for a sequence of tokens $w = (w_1, \dots, w_L)$, the loss is defined as:
\begin{equation}
\label{eq:pretrain loss}
\mathcal{L}_{\text{nll}} = - \sum_{i=1}^{L} \log P(w_i \mid w_{<i} ; \Theta) \, ,
\end{equation}
where $\Theta \in \{\theta_E, \theta_D\}$ represents the parameters of the respective modules. \emph{This phase establishes an intrinsic data discrepancy between the two modules to facilitate subsequent compression.} 
A critical methodological gap in prior work that encoders and decoders are often built on off-the-shelf pretrained LLMs whose pretraining corpora are unavailable, making it difficult to reason about the models’ endogenous data distributions. To overcome this limitation and enable a principled investigation, we pretrain a suite of LLMs \emph{from scratch}, varying both the pretraining data distribution and the parameter scale. This allows us to create bespoke encoder and decoder backbones with precisely controlled parameter counts and, most importantly, distinct endogenous data distributions.
\paragraph{Fine-tuning.} The encoder and decoder are coupled into a joint autoencoder architecture. The encoder maps the original context and learnable tokens into memory slots, while the decoder is frozen. Prompted by a specialized $[\text{AE}]$ indicator token, the decoder reconstructs the original context $c$ via teacher-forcing. We optimize only the encoder parameters to minimize the reconstruction loss. The training objective is:
\begin{equation}
\label{eq:finetune_object}
\begin{aligned}
\mathcal{L}_{\text{AE}} &= \max_{\widetilde{m}_1, \ldots, \widetilde{m}_k} P(c \mid \widetilde{m}_1, \ldots, \widetilde{m}_k ; \Theta_{E}) \\
&= \max_{e_m} P(c \mid m_1, \ldots, m_k ; \Theta_{E}, e_m) \, .
\end{aligned}
\end{equation}

To imbue these from-scratch pretrained models with context compression capability, we apply a fine-tuning procedure after pretraining. This asymmetric update strategy is designed to facilitate \textit{plug-in} compression, allowing the encoder to adapt to the fixed decoder for more generalizable and modular deployment in various scenarios. The data used for this fine-tuning stage are aligned with the encoder’s endogenous distribution. Moreover, to prevent input–output misalignment caused by mismatched model sizes between the encoder and decoder, we insert a linear projection layer to map encoder slot vectors into the decoder’s representation space, which mitigates representation mismatch and improves semantic alignment across model scales. The overall framework is illustrated in Figure~\ref{fig:framework}.

\subsection{Evaluation Metrics}
We employ \textbf{F1 score}~\cite{rajpurkar-etal-2016-squad}, \textbf{ROUGE-L}~\cite{lin-2004-rouge}, and \textbf{BLEU}~\cite{papineni-etal-2002-bleu} to measure the quality of the reconstructed text. This choice is motivated by the need to rigorously evaluate reconstruction fidelity across multiple granularities: \textbf{F1 score} provides a balanced measure of token-level precision and recall, reflecting the model's ability to recover the exact vocabulary set. \textbf{ROUGE-L} utilizes the Longest Common Subsequence (LCS) to assess structural similarity and sentence-level fluency, ensuring that the global dependencies of the input are preserved. \textbf{BLEU} quantifies $n$-gram overlap (we use 4-gram in this paper), serving as a proxy for the local consistency and precision of the generated sequences relative to the ground truth.
Together, these metrics offer a comprehensive quantitative assessment of whether the latent bottleneck effectively captures and preserves the essential semantic and structural information of the input text.

\section{Experiments}
We aim to provide a systematic analysis of the impact of data distribution on context compression including \textbf{input data} and \textbf{intrinsic data}. To guide our study, we formalize our investigation through the following Research Questions (RQs): \textbf{RQ1:} How does data distribution affect compression quality? \textbf{RQ2:} Which component’s internal prior exerts a more dominant influence? \textbf{RQ3:} How does scalability impacts performance? \textbf{RQ4:} How is the efficiency of compression and generation? \textbf{RQ5:} In the extreme case, what will happen if there is a large intrinsic data distribution gap between the encoder and the decoder?

\begin{figure}[htb]
    \centering
    \begin{subfigure}[b]{0.7\linewidth}
        \centering
        \includegraphics[width=\linewidth]{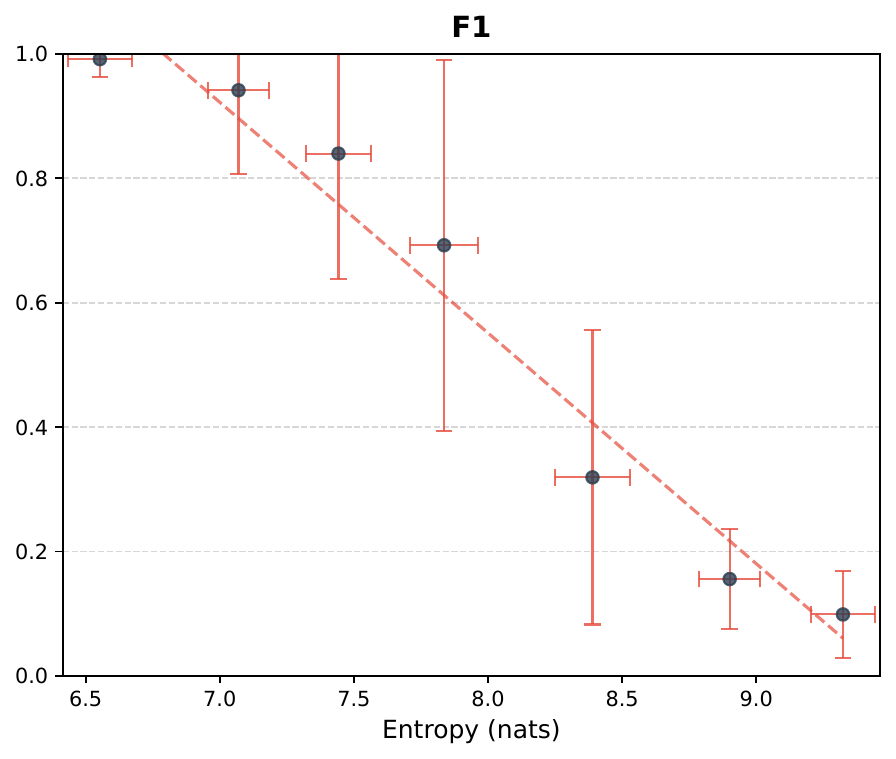}
        \caption{The relationship between entropy computed by the encoder and compression quality.}
        \label{fig:top}
    \end{subfigure}
    
    \vspace{1em} 
    
    \begin{subfigure}[b]{0.7\linewidth}
        \centering
        \includegraphics[width=\linewidth]{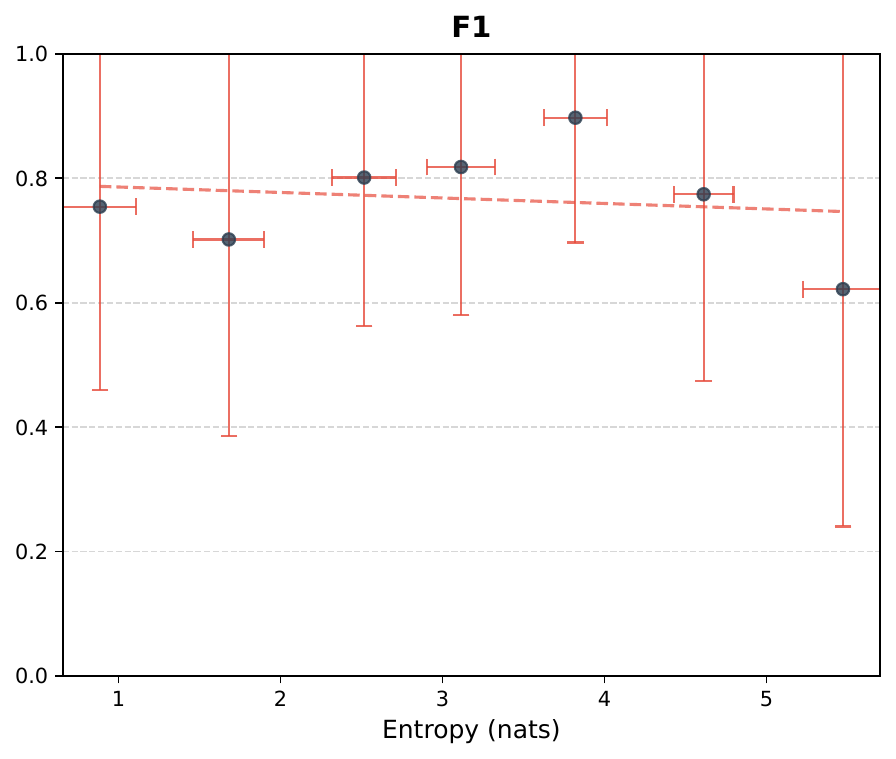}
        \caption{The relationship between entropy computed by the decoder and compression quality.}
        \label{fig:bottom}
    \end{subfigure}
    
    \caption{The impact of input data entropy on the compression process. For the encoder, input data entropy is negatively correlated with compression quality. In contrast, for the decoder, input data entropy shows no clear association with compression quality.\emph{The encoder and decoder sizes are fixed at 500M to focus on the impact of data distribution.}}
    \label{fig:input_diff}
\end{figure}

\begin{figure*}[ht]
    \centering
    \includegraphics[width=1\linewidth]{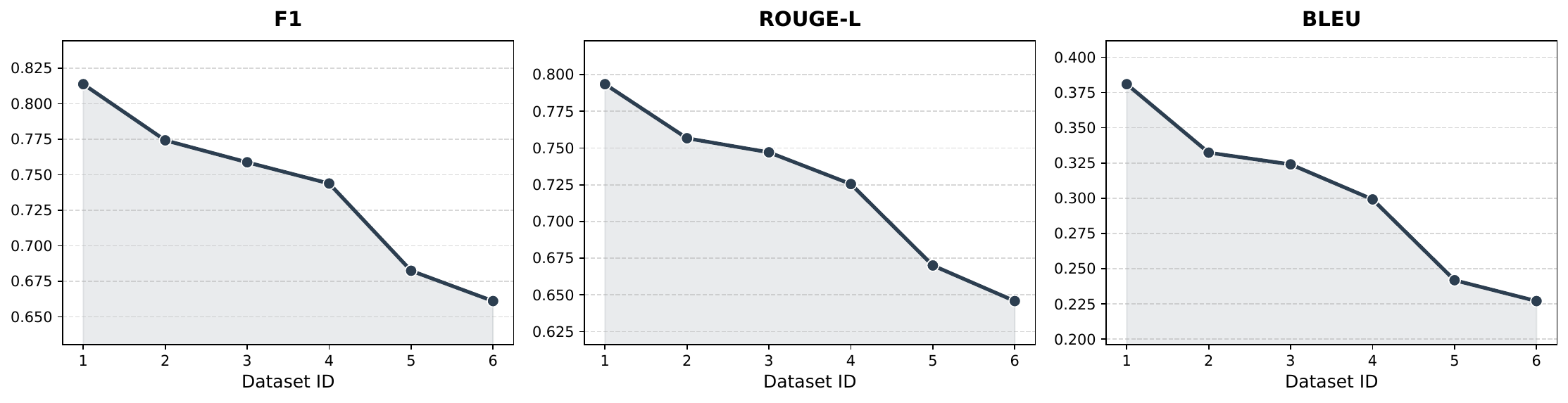}
    \caption{The impact of intrinsic data (\textit{i.e.}, pretrained knowledge) gap on the compression process. The encoder's pretraining data always comes from \textbf{dataset $\mathcal{D}_1$}, while the decoder's pretraining data comes from \textbf{dataset $\mathcal{D}_1$ to dataset $\mathcal{D}_6$}, as indicated by the horizontal axis. The test data is fixed as $\mathcal{D}_1$. As the divergence between the encoder and decoder increases, all metrics exhibit a clear downward trend. \emph{The encoder and decoder sizes are fixed at 500M to focus on the impact of data distribution.}}
    \label{fig:intrinsic_diff}
\end{figure*}

\subsection{Experimental Setup}
\paragraph{Models.} As previously mentioned, the pre-training corpus distributions of most pre-trained models are not publicly accessible. It is difficult to disentangle the effect of data distribution from the effect of latent knowledge embedded during pretraining. To obtain full control over endogenous distributions, we therefore pretrain a series of base models from scratch using the standard Qwen3~\cite{qwen2025qwen25technicalreport} architecture as encoder and decoder backbones. Specifically, we use pre-training corpora with different data distributions $\mathcal{D}_i$ to train a suite of models with varying sizes $N \in \{200\text{M},500\text{M}, 800\text{M}, 1\text{B}\}$.  This design yields a controlled set of encoder–decoder pairings where the endogenous distribution gap can be systematically adjusted by selecting encoders and decoders pretrained on matched versus mismatched corpora, enabling a direct study of how distributional mismatch impacts compression and reconstruction. Model architectural details as shown in Table \ref{tab:Model configuration}.

\begin{table}[ht]
\small
\setlength{\tabcolsep}{3pt}
\centering
\caption{Model configuration.}
\resizebox{1.0 \columnwidth}{!}{%
\begin{tabular}{l | c c c c c  c c}
\toprule
& \textbf{200M} & \textbf{500M} & \textbf{800M} & \textbf{1B} \\
\midrule
Model Size ($N$) & 201,643,776 & 533,126,144 & 799,686,144 & 1,026,191,872  \\
\midrule
Hidden Size & 768 & 1,024 & 1,536 & 1,536 \\
Intermediate Size & 2,048 & 3,072 & 3,072 & 4,096 \\
Num. of Layers & 12 & 24 & 24 & 28 \\
Num. of Heads & 8 & 16 & 16 & 16 \\
Num. of Key Value Heads & 4 & 8 & 8 & 8 \\
\bottomrule
\end{tabular}
}%
\label{tab:Model configuration}
\end{table}

\paragraph{Implementation Details.} Aftre pretraining, each encoder-decoder pair is subjected to a unified fine-tuning protocol to instill the target context compression and reconstruction capability. The training paradigm follows ICAE. To preserve the decoder’s learned distribution as a fixed reference, the decoder parameters are frozen throughout training, and only the encoder parameters and an additional linear projection layer are updated. Unless otherwise stated, we fix the compression ratio to $r=4$ and set the number of memory slots to $k=128$. To minimize experimental variance, all runs share an identical training pipeline and the same hyperparameter configuration (see Appendix~\ref{apx:hyper} for details). Most training and testing experiments are conducted on 64 NVIDIA H800 GPUs, while latency evaluation is measured on a single NVIDIA H20 GPU.

\subsection{Data Distribution Matters (RQ1)}
We emphasize that current context compression methods primarily focus on model-side improvements, while the impact of data distribution on context compression remains largely unexplored. We adopt a data-centric perspective to investigate how data distribution impacts compression quality from two angles: (1) the input data itself and (2) the intrinsic data (\textit{i.e.}, pretrained knowledge embedded in the model).

We employ an autoencoder framework to measure compression quality, as it explicitly reveals how much information is preserved in the compressed representation. To quantify differences across input data under a unified metric, we use entropy. As shown in Figure~\ref{fig:input_diff}, compression quality is negatively correlated with entropy from the encoder’s perspective, but shows no clear association with entropy from the decoder’s perspective, suggesting that encoder-computed entropy is predictive of compression performance. Regarding the impact of intrinsic data, we fix the encoder’s pretrained knowledge to that derived from \textbf{dataset $\mathcal{D}_1$}, while varying the decoder’s pretrained knowledge across \textbf{dataset~$\mathcal{D}_1$ to dataset $\mathcal{D}_6$} (with increasing gap between encoder and decoder). 

As shown in Figure~\ref{fig:intrinsic_diff}, we observe that larger distributional gaps in intrinsic data (\textit{i.e.}, pretrained knowledge) lead to worse compression quality, with consistent degradation across all evaluation metrics (F1, ROUGE-L, BLEU). We further conduct experiment under an extreme setting where the decoder is pretrained solely on \emph{entirely random text} (constructed by concatenating upper/lowercase letters and punctuation). In this case, all metrics drop to near zero; we provide a detailed analysis in Sec.~\ref{subsec:case_study}.

In summary, our findings demonstrate that data distribution, whether the input data or intrinsic data (\textit{i.e.}, the pretrained knowledge) has a significant impact on compression quality. Thus, \textbf{data distribution matters}. 

\begin{table}[t]
\small
\setlength{\tabcolsep}{3pt}
\centering
\caption{Results of models with misaligned intrinsic data distributions serving as encoder and decoder backbones, respectively. $LLM_{\mathcal{D}_i}$ denotes the model pre-trained on corpus $\mathcal{D}_i$, and $\mathcal{D}_i$ represents the model’s intrinsic data distribution. The Data column indicates the data distribution used in the fine-tuning framework and the corresponding distributions of the evaluation datasets. All metrics are reported as percentages (\%).}
\resizebox{ \columnwidth}{!}{%
\begin{tabular}{c c| c | c c c}
\toprule
\textbf{Encoder} & \textbf{Decoder} & \textbf{Data} & \textbf{F1} & \textbf{BLEU} & \textbf{ROUGE-L} \\
\midrule
$LLM_{\mathcal{D}_1}$ & $LLM_{\mathcal{D}_1}$ & $\mathcal{D}_1$ & 81.57 & 42.12 & 79.29 \\
$LLM_{\mathcal{D}_1}$ & $LLM_{\mathcal{D}_6}$ & $\mathcal{D}_1$ & 69.44 & 27.02 & 67.82 \\
$LLM_{\mathcal{D}_6}$ & $LLM_{\mathcal{D}_1}$ & $\mathcal{D}_1$ & 75.86 & 34.65 & 73.36 \\
\midrule
$LLM_{\mathcal{D}_1}$ & $LLM_{\mathcal{D}_1}$ & $\mathcal{D}_1$ & 81.57 & 42.12 & 79.29 \\
$LLM_{\mathcal{D}_1}$ & $LLM_{\mathcal{D}_2}$ & $\mathcal{D}_1$ & 77.59 & 35.95 & 75.92 \\
$LLM_{\mathcal{D}_2}$ & $LLM_{\mathcal{D}_1}$ & $\mathcal{D}_1$ & 80.02 & 39.82 & 77.71 \\
\bottomrule
\end{tabular}
}%
\label{tab:Encoder vs. Decoder.}
\end{table}

\subsection{Guideline I: Prioritize Alignment with Decoder’s Intrinsic Data Distribution (RQ2)}
\label{sec:Encoder vs. Decoder (RQ2)}
While our previous analysis establish that the fidelity of context compression is sensitive to the distributional divergence between the encoder and decoder, it remains unclear which component’s internal prior exerts a more dominant influence on the overall performance. Given a fixed encoder–decoder pair with endogenous divergence, is it more beneficial to make the compression data match the encoder’s intrinsic data distribution or the decoder’s intrinsic data distribution? In other words, when only one alignment can be satisfied, which alignment provides larger returns?

To investigate this, we configure an encoder-decoder pair using a model pre-trained on corpus $\mathcal{D}_1$ ($LLM_{\mathcal{D}_1}$) as the encoder and a model pre-trained on corpus $\mathcal{D}_6$ ($LLM_{\mathcal{D}_6}$) as the decoder. Then fine-tune them using a dataset randomly sampled from $\mathcal{D}_1$ and evaluate on an evaluation set also randomly sampled from $\mathcal{D}_1$. In this setting, the compression data are aligned with the encoder’s intrinsic data distribution but misaligned with the decoder’s intrinsic data distribution. Additionally, we swap the encoder and decoder backbones, using $LLM_{\mathcal{D}_6}$ as the encoder and $LLM_{\mathcal{D}_1}$ as the decoder, while keeping the fine-tuning and evaluation data drawn from $\mathcal{D}_1$. This configuration makes the compression data aligned with the decoder’s endogenous distribution, enabling a direct comparison. The results in Table \ref{tab:Encoder vs. Decoder.} show that compression performs best when the encoder and decoder share the same endogenous distribution. When they are mismatched, aligning the compression data with the decoder yields better performance than aligning it with the encoder, indicating that alignment with the decoder’s endogenous distribution is more critical. We replicate the same analysis using $\mathcal{D}_2$ as the reference distribution and observe the same trend.

In conclusion, our findings support two key claims: 1) Context compression achieves the best performance when the intrinsic distribution (\textit{i.e.}, pretrained knowledge) of the encoder and decoder are aligned. 2) When a distribution gap exists, aligning the decoder’s intrinsic data distribution with the data leads to better outcomes.

\begin{table*}[ht]
\small
\setlength{\tabcolsep}{3pt}
\centering
\caption{Scalibility of Encoder and Decoder. The performance of models across different scales with misaligned intrinsic data distributions. $LLM_{\mathcal{D}_i}$ denotes the model pre-trained on corpus $\mathcal{D}_i$, and $\mathcal{D}_i$ represents the model’s intrinsic data distribution. The Data column indicates the data distribution used in the unified fine-tuning framework and the corresponding distributions of the evaluation datasets. We \textbf{bold} the best results and \underline{underline} the second best. All metrics are reported as percentages (\%).}
\resizebox{0.7\textwidth}{!}{%
\begin{tabular}{>{\centering\arraybackslash}p{2.8cm}|>{\centering\arraybackslash}p{2.8cm}|>{\centering\arraybackslash}p{1.8cm}|>{\centering\arraybackslash}p{1.7cm}>{\centering\arraybackslash}p{1.7cm}>{\centering\arraybackslash}p{2.05cm}}
\toprule
\textbf{Encoder} & \textbf{Decoder} & \textbf{Data} & \textbf{F1} & \textbf{BLEU} & \textbf{ROUGE-L} \\
\midrule
$LLM_{\mathcal{D}_1}$(200M) & \multirow{4}{*}{$LLM_{\mathcal{D}_6}$(500M)} & \multirow{4}{*}{$\mathcal{D}_1$} & 65.40 & 22.30 & 63.31 \\
$LLM_{\mathcal{D}_1}$(500M) & &  & 69.44 & 27.02 & 67.82 \\
$LLM_{\mathcal{D}_1}$(800M) & &  & 72.43 & 30.34 & 70.94 \\
$LLM_{\mathcal{D}_1}$(1B) & &  & 72.62 & 30.80 & 70.48 \\
\midrule
\multirow{4}{*}{$LLM_{\mathcal{D}_6}$(500M)} & $LLM_{\mathcal{D}_1}$(200M) & \multirow{4}{*}{$\mathcal{D}_1$} & 68.09 & 26.73 & 65.45 \\
& $LLM_{\mathcal{D}_1}$(500M) &  & 75.86 & 34.65 & 73.36 \\
& $LLM_{\mathcal{D}_1}$(800M) &  & 79.01 & 37.08 & 77.83 \\
& $LLM_{\mathcal{D}_1}$(1B) &  & \textbf{82.87} & \textbf{45.08} & \textbf{81.21} \\
\midrule
\cellcolor{lightgray}$LLM_{\mathcal{D}_1}$(500M) & \cellcolor{lightgray}$LLM_{\mathcal{D}_1}$(500M) & \cellcolor{lightgray}$\mathcal{D}_1$ & \underline{\cellcolor{lightgray}81.57} & \underline{\cellcolor{lightgray}42.12} & \underline{\cellcolor{lightgray}79.29} \\
\midrule
$LLM_{\mathcal{D}_1}$(200M) & \multirow{4}{*}{$LLM_{\mathcal{D}_6}$(500M)} & \multirow{4}{*}{$\mathcal{D}_6$} & 40.59 & 8.09 & 37.46 \\
$LLM_{\mathcal{D}_1}$(500M) & &  & 46.28 & 9.66 & 45.48 \\
$LLM_{\mathcal{D}_1}$(800M) & &  & 50.35 & 11.50 & 49.03 \\
$LLM_{\mathcal{D}_1}$(1B) & &  & \textbf{58.50} & \textbf{16.16} & \textbf{56.93} \\
\midrule
\multirow{4}{*}{$LLM_{\mathcal{D}_6}$(500M)} & $LLM_{\mathcal{D}_1}$(200M) & \multirow{4}{*}{$\mathcal{D}_6$} & 38.01 & 6.89 & 37.66 \\
& $LLM_{\mathcal{D}_1}$(500M) &  & 47.74 & 13.37 & 46.47 \\
& $LLM_{\mathcal{D}_1}$(800M) &  & 45.04 & 8.80 & 45.13 \\
& $LLM_{\mathcal{D}_1}$(1B) &  & 42.96 & 7.85 & 43.07 \\
\midrule
\cellcolor{lightgray}$LLM_{\mathcal{D}_6}$(500M) & \cellcolor{lightgray}$LLM_{\mathcal{D}_6}$(500M) & \cellcolor{lightgray}$\mathcal{D}_6$ & \underline{\cellcolor{lightgray}56.24} & \underline{\cellcolor{lightgray}14.18} & \underline{\cellcolor{lightgray}55.97} \\
\bottomrule
\end{tabular}
}
\label{tab:Scalibility}
\end{table*}

\subsection{Guideline II: Prioritize Distribution Alignment Over Pure Scaling for Compute Efficiency (RQ3)}
The remarkable success of LLMs is often attributed to scaling laws~\cite{kaplan2020scalinglawsneurallanguage,10.5555/3600270.3602446}, where performance reliably improves with model size. This paradigm, however, raises a critical question in our context: can the performance degradation caused by distributional mismatch be mitigated simply by scaling up the model? There are two competing hypotheses. A larger encoder may act as a stronger “distribution translator,” producing latents that better capture salient information from the source domain and are more readily consumable by an out-of-domain decoder. Alternatively, a larger decoder may be more robust to mismatch and information bottlenecks, exploiting stronger priors to infer omitted details and improve reconstruction from imperfect compressed signals. 

To systematically dissect these, we conducted a systematic study on the scalability of the encoder and decoder. Specifically, we trained a series of models on corpus $\mathcal{D}_1$ with varying sizes ($N \in \{200\text{M},500\text{M}, 800\text{M}, 1\text{B}\}$). These models were then paired with a fixed $LLM_{\mathcal{D}_6}$(500M) backbone to create two sets of experiments under a controlled distributional gap, one scaling the encoder and the other scaling the decoder. Afterward, we fine-tune and evaluate these encoder–decoder pairs using fine-tuning data drawn from $\mathcal{D}_1$ and $\mathcal{D}_6$ respectively.

\begin{figure}[ht]
\begin{center}
\centerline{\includegraphics[width=\columnwidth]{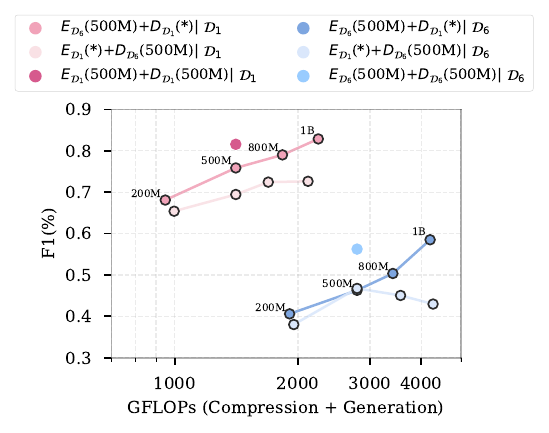}}
\caption{FLOPs comparison across various encoder-decoder size combinations. The y-axis reports the F1 score, and the x-axis shows the average compute cost in FLOPs per generation (log scale). $E_{\mathcal{D}_i}$(500M)+$D_{\mathcal{D}_i}(*)|$  $\mathcal{D}_i$ indicates that the encoder backbone is fixed to $LLM_{\mathcal{D}_i}$(500M) while the decoder size is varied (and vice versa). $|\mathcal{D}_i$ denotes the data distribution used in the unified fine-tuning framework and for evaluation. Because different distributions contain different numbers of tokens, the computed FLOPs will vary accordingly.}
\label{fig:flops}
\end{center}
\vspace{-2.5em}
\end{figure}

\begin{figure}[ht]
\begin{center}
\centerline{\includegraphics[width=\columnwidth]{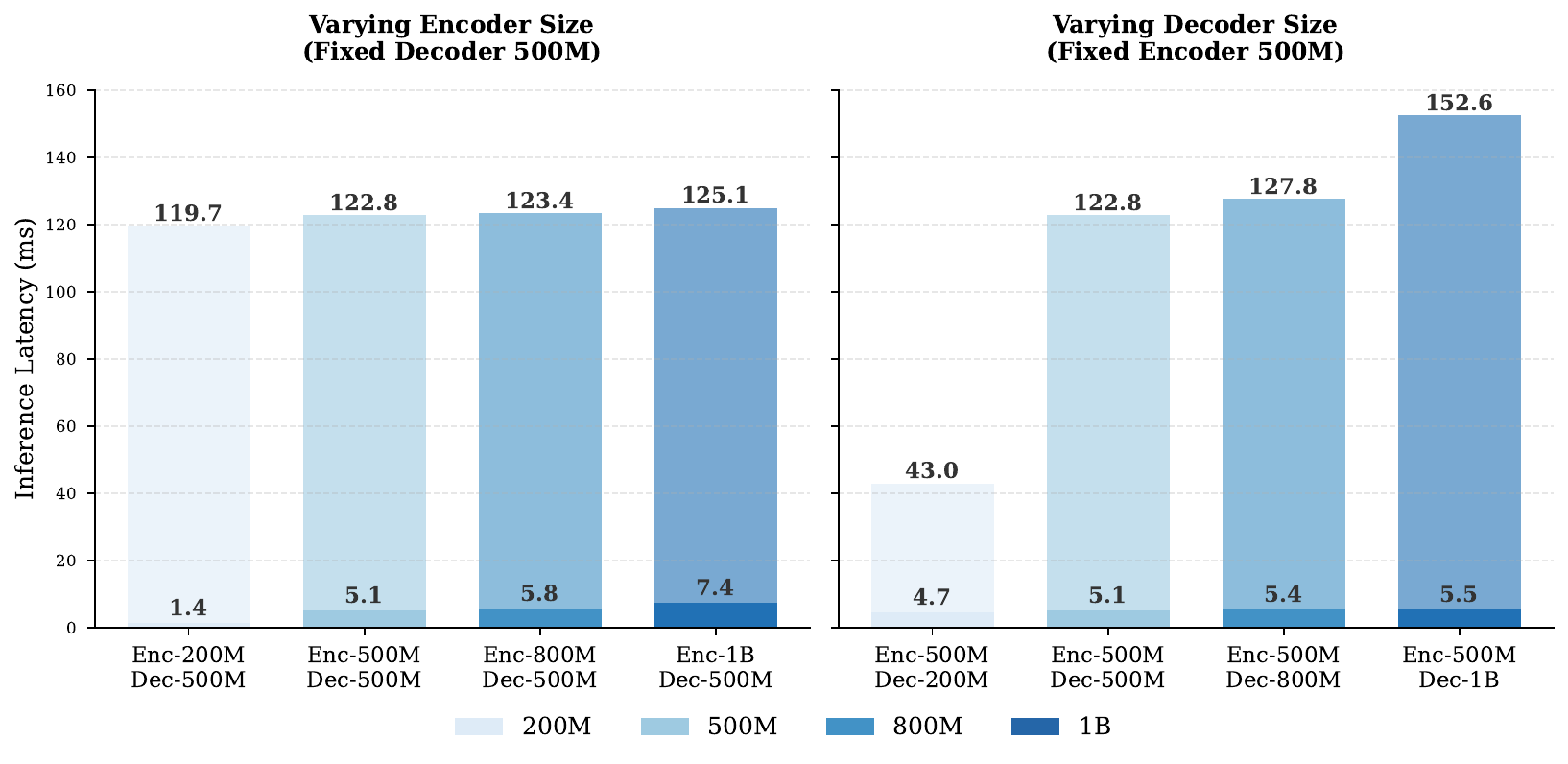}}
\caption{Inference efficiency analysis. We compare the inference efficiency of encoder–decoder combinations with different model sizes under a fixed context length of 4096. In the stacked bars, the lower segment denotes compression latency, and the upper segment denotes generation latency.}
\label{fig:inference_latency}
\end{center}
\vspace{-1.5em}
\end{figure}

As shown in Table \ref{tab:Scalibility}, when the decoder is fixed and the compression data distribution is misaligned with the decoder’s endogenous distribution, increasing the encoder size improves compression quality, but the marginal gains diminish as the encoder becomes larger. In contrast, when the encoder is fixed, increasing the decoder size yields larger improvements than enlarging the encoder under the previous setting. This further supports the conclusion in Section \ref{sec:Encoder vs. Decoder (RQ2)} that alignment with the decoder’s endogenous distribution is more critical.  We additionally conduct experiment under distributional alignment, where the encoder and decoder are pretrained on the same distribution and the compression context is also aligned. In this case, even smaller models can achieve performance comparable to, or better than, larger models under mismatch. Concretely, the aligned 500M–500M pair configuration outperforms the 500M–800M pair configuration and even approaches the performance of the 500M–1B pair configuration. We replicate the same experiment using $\mathcal{D}_6$ as the reference distribution and observe the same qualitative trend.

These results indicate that increasing parameter count can partially mitigate mismatch-induced degradation, but it is not a substitute for distributional alignment. Data alignment should be considered a primary optimization target, offering a more parameter-efficient path to better performance.

\subsection{Efficiency Analysis (RQ4)}
In many existing soft-prompt approaches~\cite{ge2024incontext,cao-etal-2024-retaining,he2025informationtheoreticperspectiveagentic}, the encoder is typically almost as large as the decoder, whose compute cost can be comparable to that of the decoder. As a result, the computation time spent on compression can be comparable to the time the original LLM needs to process the input~\cite{li-etal-2025-prompt}. This reduces the practical efficiency benefits of compression and introduces a practical trade-off: how should compute and parameters be allocated between the encoder and decoder to maximize the overall return of compression? To study this, we compute the FLOPs-per-generation spend for the configurations (Table \ref{tab:Scalibility}) and plot the results in Figure \ref{fig:flops}. Concurrently, we conduct empirical tests to evaluate the actual inference latency, with the findings presented in Figure \ref{fig:inference_latency}. Synthesizing these results, it is suggested that the optimal compute allocation favors a larger decoder over a complex encoder.

\begin{table}[ht] 
\centering
\caption{A case study where the encoder's pretrained knowledge originates from $\mathcal{D}_1$, while the decoder's pretrained knowledge consists of \emph{entirely random data} (\emph{i.e.}, each sample is a random combination of uppercase and lowercase letters and punctuation marks).}\label{tab:ae_examples}
\small
\scalebox{0.93}{
\begin{tabular}{p{\columnwidth}} 
\toprule
\textbf{Origin Context} \\ \midrule
FAQs FAQs How long does it take to hear about my grant application? This year, we are open to applications from 30 May – 30 June 2017, giving you a month to apply. During this time we will be processing your applications and not making any awards. We may contact you during this time if we need any additional information. We will contact you by letter with our decision on your application by 28 July. I’ve applied to Take a Break before and got a grant, can I apply again? Yes, you can make one application per year. I have two disabled children at home – can I make two Take a Break applications? We can make only one award per household in each funding round. While we can sometimes make a larger award, this is not based on the number of disabled children or children in a household but on exceptional circumstances. I have forgotten my Take a Break number. Don’t worry, as long as the rest of your application form is completed we will do the rest. \\ \midrule

\textbf{Reconstruction Context} \\ \midrule
\cellcolor{yellow!30}
\seqsplit{RgggggggggggggkgkgkgkRg0g5g5g5g3g3g3g5g1g3g0g0g1g0g1g1g2g3g1g3g0g1g3g0g0g3g3g3g3g3g1g3g3g3g0g5g2g3g3g3g3g3g3g1g1g1g1g3sg1g2ig0g2iRbsg1g0g0g0g1g0g3rg0Z4F1Rbqrg3rsg3rsg0sg3rsg0rj5rj2rj2rj2rj2rj2rj2rj2rj20rj2r} \\ \bottomrule
\end{tabular}
}
\end{table}

\subsection{Case Study (RQ5)}
\label{subsec:case_study}
We consider an extreme case where the encoder is pre-trained on the corpus $\mathcal{D}_1$, while the decoder is pre-trained on a synthetic corpus of random text (where each sample consists of a stochastic combination of punctuation marks and alphanumeric characters). As illustrated in Table~\ref{tab:ae_examples}, the decoder in this scenario generates outputs that are entirely random, mirroring the noise in its pre-training data. This observation suggests that a substantial discrepancy between the distributions of intrinsic data leads to a drastic degradation in compression quality, or even complete failure.


\section{Conclusion}
In this paper, we present the first systematic, data-centric study of context compression for Large Language Models (LLMs). Using an autoencoder framework trained from scratch across diverse data distributions, we disentangle how input complexity and model priors affect compression quality. Our experiments reveal three key insights: (1) encoder-measured input entropy reliably predicts worse compression, while the frozen decoder’s complexity judgment is biased by its internal priors; (2) mismatched data distributions between encoder and decoder impair information preservation; and (3) the decoder’s intrinsic distribution dominates performance more than the encoder’s, highlighting a fundamental asymmetry.

We thus propose two practical guidelines: align encoder data with the decoder’s distribution rather than scaling parameters indiscriminately, and prioritize a larger decoder over a complex encoder for better compute efficiency. We hope this work redirects context compression research toward understanding data-distribution effects.








\section*{Impact Statement}
This paper presents a systematic, data-centric investigation into context compression for Large Language Models (LLMs). It analyzes how input data complexity (entropy) and the intrinsic knowledge gap between encoders and decoders influence compression quality, providing practical guidelines for optimal computational resource allocation. The data and models used in this work are sourced from publicly available benchmarks and open-source platforms under appropriate licenses. While our findings may influence the design and deployment of efficient long-context LLM systems, they do not introduce new ethical risks beyond those already present in existing context compression and language modeling research. Thus, no additional ethical concerns require specific attention.

\nocite{langley00}

\bibliography{example_paper}
\bibliographystyle{icml2026}

\newpage
\appendix
\onecolumn
\section{The Pile Dataset}
\label{apx:pile}

\textbf{The Pile}~\cite{gao2020pile} is a large-scale, diverse English text dataset comprising 825.18 GiB of high-quality data designed for language modeling. Developed by EleutherAI, it consists of 22 distinct sub-datasets that emphasize cross-domain generalization and high-quality academic or professional sources. 

The components of The Pile are categorized into several domains including academic writing, books, web content, code, and legal documents. The full list of its 22 sub-categories is as follows:

\begin{itemize}
    \item \textbf{Pile-CC}: Filtered and extracted Common Crawl (CC) data.
    \item \textbf{PubMed Central}: Full-text biomedical research articles.
    \item \textbf{Books3}: A large-scale collection of books from the Bibliotik tracker.
    \item \textbf{OpenWebText2}: Enhanced scrape of Reddit outgoing links.
    \item \textbf{ArXiv}: LaTeX-based scientific preprints in technical fields.
    \item \textbf{GitHub}: Open-source code repositories.
    \item \textbf{FreeLaw}: Legal opinions from US federal and state courts.
    \item \textbf{Stack Exchange}: Question-answer pairs across diverse topics.
    \item \textbf{USPTO Backgrounds}: Technical backgrounds from US patents.
    \item \textbf{PubMed Abstracts}: Summaries of over 30 million biomedical publications.
    \item \textbf{Project Gutenberg (PG-19)}: Classic Western literature.
    \item \textbf{OpenSubtitles}: Movie and television dialogue transcripts.
    \item \textbf{Wikipedia (en)}: Comprehensive encyclopedia articles.
    \item \textbf{DM Mathematics}: Mathematical problems and reasoning tasks.
    \item \textbf{Ubuntu IRC}: Spontaneous human interaction from chat logs.
    \item \textbf{BookCorpus2}: Extended version of the original BookCorpus.
    \item \textbf{EuroParl}: Proceedings of the European Parliament.
    \item \textbf{HackerNews}: High-quality intellectual dialogue from community comments.
    \item \textbf{YouTube Subtitles}: Manually generated video captions.
    \item \textbf{PhilPapers}: Academic philosophy publications.
    \item \textbf{NIH ExPorter}: Scientific abstracts of awarded research grants.
    \item \textbf{Enron Emails}: Real-world professional email communications.
\end{itemize}
The foundational dataset, $\mathcal{D}_1$, is composed entirely of the Pile-CC subset. For datasets $\mathcal{D}_2$ through $\mathcal{D}_6$, we progressively substitute the content with a mixture of ArXiv, GitHub, and DM Mathematics in a ratio of $2:2:1$, scaling the integration proportions from $1/6$ to $5/6$. \textit{This experimental design is motivated by the need to investigate the divergence in compression performance between general natural language and formal logical corpora. While natural language is typically associated with information-seeking tasks, code and mathematics are emblematic of logical reasoning, representing two primary and functionally distinct categories of textual data in language modeling.}

\section{Theoretical Analysis}
\label{apx:theoretical_analysis}
We provide an information-theoretic analysis of our empirical observations by first formalizing the learning objective and deriving a general decomposition of the reconstruction loss.

Let $X\sim p_{\text{data}}$ be the input context, $Z=E(X)\in\mathcal{Z}$ be the compressed code (fixed $k$ memory slots), and $X'$ be the reconstruction sampled from the frozen decoder $p_D(\cdot\mid Z)$. Training optimizes only the encoder:
\begin{equation}
\label{eq:ae_obj}
\inf_{\theta_E}\ \mathcal{L}_{\text{AE}}(\theta_E)
\triangleq
\inf_{\theta_E}\ \mathbb{E}_{X\sim p_{\text{data}}}\big[-\log p_D(X\mid E(X;\theta_E))\big].
\end{equation}
Define the decoder-reachable conditional family under the compression budget
\begin{equation}
\label{eq:F_D}
\mathcal{F}_D \triangleq \{p_D(\cdot\mid z): z\in\mathcal{Z}\}.
\end{equation}
For any fixed encoder, define $q_E(\cdot)\triangleq p_D(\cdot\mid E(\cdot))\in\mathcal{F}_D$, thus by the cross-entropy decomposition,
\begin{equation}
\label{eq:lower_bound_family}
\inf_{\theta_E}\mathcal{L}_{\text{AE}}
=
H(p_{\text{data}})+\inf_{q\in\mathcal{F}_D}D_{\mathrm{KL}}(p_{\text{data}}\|q)
\ \ge\
H(p_{\text{data}}),
\end{equation}
where $H(p_{\text{data}})\triangleq \mathbb{E}_{X\sim p_{\text{data}}}[-\log p_{\text{data}}(X)]$.
Eq.~\eqref{eq:lower_bound_family} makes explicit two sources of reconstruction error: (i) an \emph{intrinsic complexity} term $H(p_{\text{data}})$ and (ii) a \emph{distribution mismatch} term determined by how close $p_{\text{data}}$ is to $q\in\mathcal{F}_D$ under a fixed compression budget.

\subsection{Why higher-entropy inputs are harder (RQ1).}
The compression budget induces an effective rate constraint. Rate--distortion theory~\cite{cover1999elements} states that achieving expected distortion $D$ requires at least
\begin{equation}
\label{eq:rd}
R(D)=\min_{p(X'|X):\ \mathbb{E}[d(X,X')]\le D} I(X;X'),
\end{equation}
and for discrete lossless reconstruction ($D=0$),
\begin{equation}
\label{eq:r0}
R(0)=H(X).
\end{equation}
Under a fixed budget (fixed effective rate $R$), if $H(X)>R$ then lossless reconstruction is impossible.
More generally, higher-entropy sources typically require a larger rate to achieve comparable fidelity, so a fixed budget leads to worse reconstruction on higher-entropy inputs.

\subsection{Intrinsic data gap as mismatch to the decoder (RQ1, RQ4) and why decoder alignment dominates (RQ2).}
Let $p_E$ and $p_D$ denote the intrinsic (pretraining-induced) data distributions of the encoder/decoder. When the gap between $p_E$ and $p_D$ is larger, $p_{\text{data}}$ aligned with $p_E$ is typically farther from the reachable conditional family of the decoder trained in $p_D$, which manifests as a larger mismatch term in Eq.~\eqref{eq:lower_bound_family}.

Crucially, aligning the \emph{decoder} with $p_{\text{data}}$ directly reduces the mismatch term, while aligning only the encoder does not efficiently change $\mathcal{F}_D$ and therefore cannot remove the distributional gap. This conforms to the observed asymmetry: decoder alignment is more important than encoder alignment.

\subsection{Scaling vs.\ alignment (RQ3).}
Increasing model size (especially the decoder) can reduce the mismatch term in
Eq.~\eqref{eq:lower_bound_family} by increasing the capacity of $\mathcal{F}_D$.
However, when the intrinsic gap is large, this term remains substantial under the fixed compression budget, yielding diminishing returns from further scaling.
In contrast, distribution alignment reduces the mismatch error more directly by moving $p_{\text{data}}$ closer to $\mathcal{F}_D$, which is more parameter-efficient in our experiments.

\section{Hyperparameters of Training}
\label{apx:hyper}
The hyperparameters for pre-training and fine-tuning are listed in Table~\ref{tab:pre_hyper_para} and Table~\ref{tab:ft_hyper_para}, respectively.

\begin{table}
    \centering
    \begin{minipage}{0.48\textwidth}
        \centering
        \caption{The hyperparameters of pretraining.}
        \label{tab:pre_hyper_para}
        \begin{tabular}{lc}
            \toprule
            \textbf{Hyperparameters} & \textbf{Value}\\
            \midrule
            Warm-up Steps & 2000  \\
            Training epochs & 1 \\ 
            Gradient Accumulation Steps & 4 \\
            Train Batch Size Per Device & 512 \\
            Max Sequence Length & 8192 \\
            Learning Rate Scheduler & cosine \\
            Max Learning Rate & 3e-4 \\
            Min Learning Rate & 3e-5 \\
            Numbers of GPUs & 128 \\
            \bottomrule
        \end{tabular}
    \end{minipage}
    \hfill 
    \begin{minipage}{0.48\textwidth}
        \centering
        \caption{The hyperparameters of fine-tuning.}
        \label{tab:ft_hyper_para}
        \begin{tabular}{lc}
            \toprule
            \textbf{Hyperparameters} & \textbf{Value}\\
            \midrule
            Warm-up Steps & 100  \\ 
            Training epochs & 3 \\ 
            Gradient Accumulation Steps & 1 \\
            Train Batch Size Per Device & 32 \\
            Weight decay & 0.1 \\
            Learning Rate Scheduler & linear \\
            Max Learning Rate & 1e-4 \\
            Min Learning Rate & 1e-5 \\
            Numbers of GPUs & 32 \\
            \bottomrule
        \end{tabular}
    \end{minipage}
\end{table}



\end{document}